\documentclass[runningheads,Anonymous]{llncs}
\usepackage[T1]{fontenc}
\usepackage{graphicx}
\usepackage{color}
\usepackage{amsmath}
\usepackage{multirow}
\usepackage{amsfonts}
\usepackage{subfigure}
\usepackage{tikz}
\usepackage{pgfplots}
\usepackage{pgfplotstable}
\usepackage{makecell}

\begin{document}
\title{Generative Sentiment Transfer via Adaptive Masking}

\author{
Yingze Xie\inst{1}, Jie Xu\inst{1}, LiQiang Qiao\inst{1}, Yun Liu\inst{2}, Feiren Huang\inst{3}, Chaozhuo Li\inst{4}
}
\authorrunning{Yingze Xie et al.}
%
\institute{School of Information Science and Technology, Beijing Foreign Studies University, Beijing, China \and
Moutai Institute, China \and
School of Information Science and Technology, Jinan University, China \and
Microsoft Research Asia,  Beijing, China
}
\maketitle              
\vspace{-6mm}
\begin{abstract} 
Sentiment transfer aims at revising the input text to satisfy a given sentiment polarity while retaining the original semantic content. The nucleus of sentiment transfer lies in precisely separating the sentiment information from the content information. Existing explicit approaches generally identify and mask sentiment tokens simply based on prior linguistic knowledge and manually-defined rules, leading to low generality and undesirable transfer performance. In this paper, we view the positions to be masked as the learnable parameters, and further propose a novel AM-ST model to learn adaptive task-relevant masks based on the attention mechanism. Moreover, a sentiment-aware masked language model is further proposed to fill in the blanks in the masked positions by incorporating both context and sentiment polarity to capture the multi-grained semantics comprehensively.  AM-ST is thoroughly evaluated on two popular datasets, and the experimental results demonstrate the superiority of our proposal. 

\end{abstract}
\section{Introduction}
\noindent Sentiment transfer~\cite{Amazon}  aims at altering the sentiment polarity of a text while preserving its vanilla content meanings, which has been widely employed in a myriad of applications such as  news sentiment transformation~\cite{fu2018} , passage editing~\cite{pessageEditing} and data augmentation~\cite{DataAugmentation}.  
For example, when inputting a text sequence ``stale food and poor service'', the expected sentiment-transferred output would be ``fresh food and good service'', which modifies the sentiment polarity from negative to positive while maintaining  the content information. 



Existing sentiment transfer models could be roughly categorized into two categories. 
The first type of work implicitly disentangles content and sentiment \cite{john2018,shen2017,fu2018,controlledTextGneration,liu2022semi,krishna2022few,chen2018adversarial} by learning the latent representations of content and sentiment respectively, and then combining the content representation and target sentiment signal to generate transferred sentences. 
Such implicit approaches generally employ GAN (Generative Adversarial Network) to remove sentiment attributes from content representation and generate text indistinguishable from real data. 
However, previous works \cite{hu2022review} noted that implicit methods generally suffer from inferior performance on content preservation and low interpretability.  
Another way of sentiment transfer is decoupling content and sentiment explicitly \cite{li2018delete,wu2019,pseudo-parallelCorpora,sudhakar2019TransformingExplicit,zhang2018learning,madaan2020politeness,guu2018generating,lee2020stable}, which first identifies sentiment-associated words and replaces them with words related to the target sentiment while keeping other words unchanged. 
Explicit methods benefit from the simplicity and explainability since the sentiment-tokens are explicitly uncovered and replaced.

Despite the promising performance of existing sentiment transfer models, they are still facing two crucial challenges. 
First, it is intractable to separate the sentiment style from the semantic content precisely. 
Existing implicit methods generally split the hidden representation of the input sequence into two vectors, which contain style and content information, respectively. 
However, such implicit methods generally perform unsatisfactorily on content preservation subtask~\cite{hu2022review}, probably brought by the loss of content information in the process of disentangling. 
Explicit models are capable of identifying emotion-associated tokens explicitly, while they generally locate these tokens simply based on prior linguistic knowledge and rules. 
Such heuristic methods are incapable of ensuring the precise correlations between tokens in masking positions and sentiment signals, leading to an obscure disentanglement. 
Second, it is nontrivial to combine content and target sentiment to generate a target sentence effectively. 
Existing works usually assume that the overall emotional tendency of the generated sentence should be close to the target sentiment label. 
\cite{wu2019} leverages Attribute Conditional Masked Language Model(AC-MLM) to fill in masked positions.  \cite{li2018delete} retrieves new phrases associated with target attributes from the corpus and combines them with content information. 
However, these methods only focus on sentence-level sentiment labels and pay no attention to the word-level polarities, thus cannot capture such fine-grained semantic information and the connections between sentence-level sentiment and word-level polarity.


In this paper, we propose a novel \textbf{S}entiment \textbf{T}ransfer model AM-ST to handle the mentioned challenges based on \textbf{A}daptive \textbf{M}asking. First, we identify emotion-associated tokens and mask them using a trainable mask module, which is capable of adaptively learning the optimal mask positions. Then,  a sentiment-aware masked language model is leveraged to fill in blanks in these masked positions, incorporating both context and sentiment polarity to improve transfer accuracy. 
Specifically, following the assumption of \cite{wu2019} that transferred sentence can be generated by simply replacing several emotional-related words, AM-ST adopts a mask classifier to identify the appropriate mask positions and then mask them with special tokens. 
In order to certify that sentiment information only appears in the masked tokens and not in the rest tokens, we design two types of losses: classification losses and adversarial losses, ensuring the clear separation of sentiment and content. 
After that, in the filling blanks phase, we adopt a sentiment-aware masked language model based on a reconstruction loss to predict both sentence- and token-level polarities in the masked positions. 
Experimental results on two popular datasets demonstrate the superiority of our proposal. 
\vspace{-1mm}
Our major contributions are summarized as follows:
\begin{itemize}
\item To alleviate the challenge of unclear disentanglement brought by low identifying quality, we propose an adaptive masking module, setting mask position as a trainable parameter to certify accuracy in identifying sentiment words. 
\item We further propose a sentiment-aware masked language model in the infilling blanks stage, which captures both semantic context and emotional signals.

\item Experimental results on popular datasets indicate that the proposed AM-ST model consistently outperforms SOTA models.
\end{itemize}




\vspace{-2mm}
\section{Related Work}
\vspace{-2mm}
Sentiment transfer is closely related to text style transfer, where the key is to modify the style of input text and retain the content. To address the problem of separating style and content, there are two major ways: 1) \textbf{Implicitly Separating:} These models divide the hidden representation of input text into two representations,  style embedding and content embedding. Fu et al.~\cite{fu2018} use adversarial networks to learn separated content representations and style representations. \cite{pseudo-parallelCorpora} generates pseudo-parallel corpora for text sentiment transfer. Yang et al.~\cite{yang2018} replace the style discriminator in the adversarial learning with a language model. 
However, \cite{lample2018} experimentally proves that simply adopting GAN is inadequate to separate style from content, and it is easy to recover style information using content representation. 
2) \textbf{Explicitly Separating:} This type of model explicitly identifies and replaces style-associated tokens. \cite{li2018delete} removes original attribute phrases and retrieves new terms related to the target attribute. \cite{wu2019} converts the emotion transfer problem into a text fill-in-the-blank task through a pre-trained masked language model. \cite{controlledTextGneration}~combines a pre-trained language model with attribute classifiers to guide text generation. Explicit methods benefit from their simplicity and explainability, because they clearly show which tokens are related to style. However, most current works exploit only prior language knowledge and rules to identify sentiment tokens without examining the association between masked tokens and sentiment information. 
Different from existing works, we adopt a trainable masked classification module to detect and mask the sentiment-related tokens. 

\vspace{-2mm}
\section{Problem Definition} 
\vspace{-2mm}

Given an input text sequence $X=\left \{ x_{1},x_{2},...,x_{N} \right \}$ ($x_{i}$ indicates a single word), its source sentiment label $l$, and the target sentiment label $\widetilde{l}$, sentiment transfer aims to generate the target sentence $\widetilde{X}$ which maintains the content information of $X$ while having the target sentiment label $\widetilde{l}$ .
Following the previous work \cite{john2018}, we select the binary sentiment label set $\left \{ positive,negative \right \}$. 
\vspace{-3mm}

\section{Methodology}
\vspace{-2mm}
\subsection{Framework}
\begin{figure*}[t]
\centering
\includegraphics[width=0.95\textwidth]{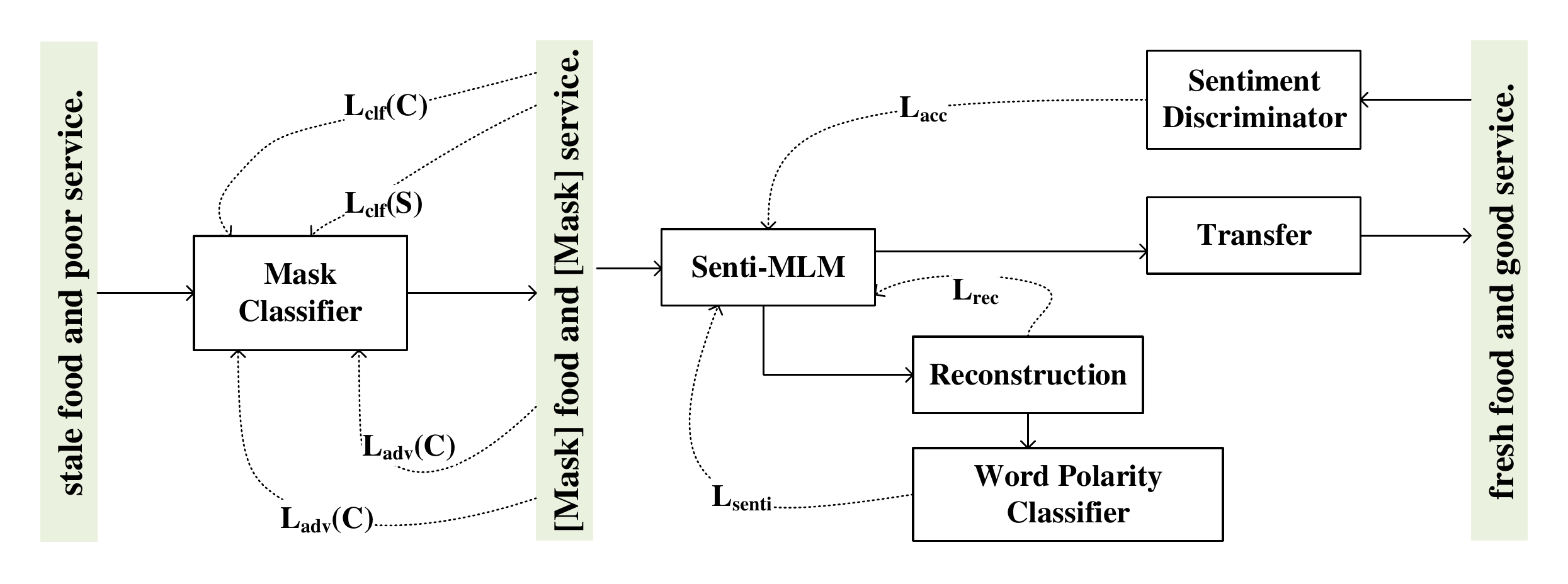} 
\vspace{-3mm}
\caption{Framework of the proposed AM-ST model. }
\label{fig1}
\vspace{-4mm}
\end{figure*}
\vspace{-2mm}
Figure \ref{fig1} demonstrates the framework of the proposed AM-ST model.  
Given the input sentence \emph{stale food and poor service}, AM-ST first utilizes the adaptive masking module to separate the sentiment words from other content words, resulting the masked sentiment token set $S$ and the content token set $C$. 
After masking the sentiment tokens with the special tokens, the sentence is converted to \emph{[mask] food and [mask] service}. 
Then, based on the learned content text $C$ and the target sentiment label $\widetilde{l}$, we further propose a generative module to predict tokens to infill words conforming target sentiment.  
Both the word-level and sentence-level sentiment polarities are properly incorporated to generate desirable sentiment-transferred sentences.  
The input text is transferred into the positive sentiment \emph{``fresh food and good service''}.  

\vspace{-2mm}
\subsection{Adaptive Sentiment Token Masking}
\vspace{-1mm}
In this stage, we aim to identify and mask sentiment-associated tokens to achieve pure content text. 
Previous works generally rely on prior linguistic knowledge (e.g., sentiment dictionary, co-occurrence frequency of words with a particular sentiment) to identify the sentiment tokens. 
However, such rule-based heuristic methods might not be optimal. 
Downstream datasets might contain unique sentiment tokens, which might be inconsistent with the general knowledge and ignored by previous works, leading to low generality and inferior performance.  
Instead of directly leveraging prior linguistic knowledge to discover and mask sentiment tokens, we view the mask positions as the learnable variables. 
Our model is expected to automatically locate the positions of the sentiment words to better align with the domain-specific knowledge. 

\begin{figure*}[t] 
\subfigure[Adaptive Sentiment Token Masking]{
\label{fig2a} 
\includegraphics[width=0.51\linewidth]{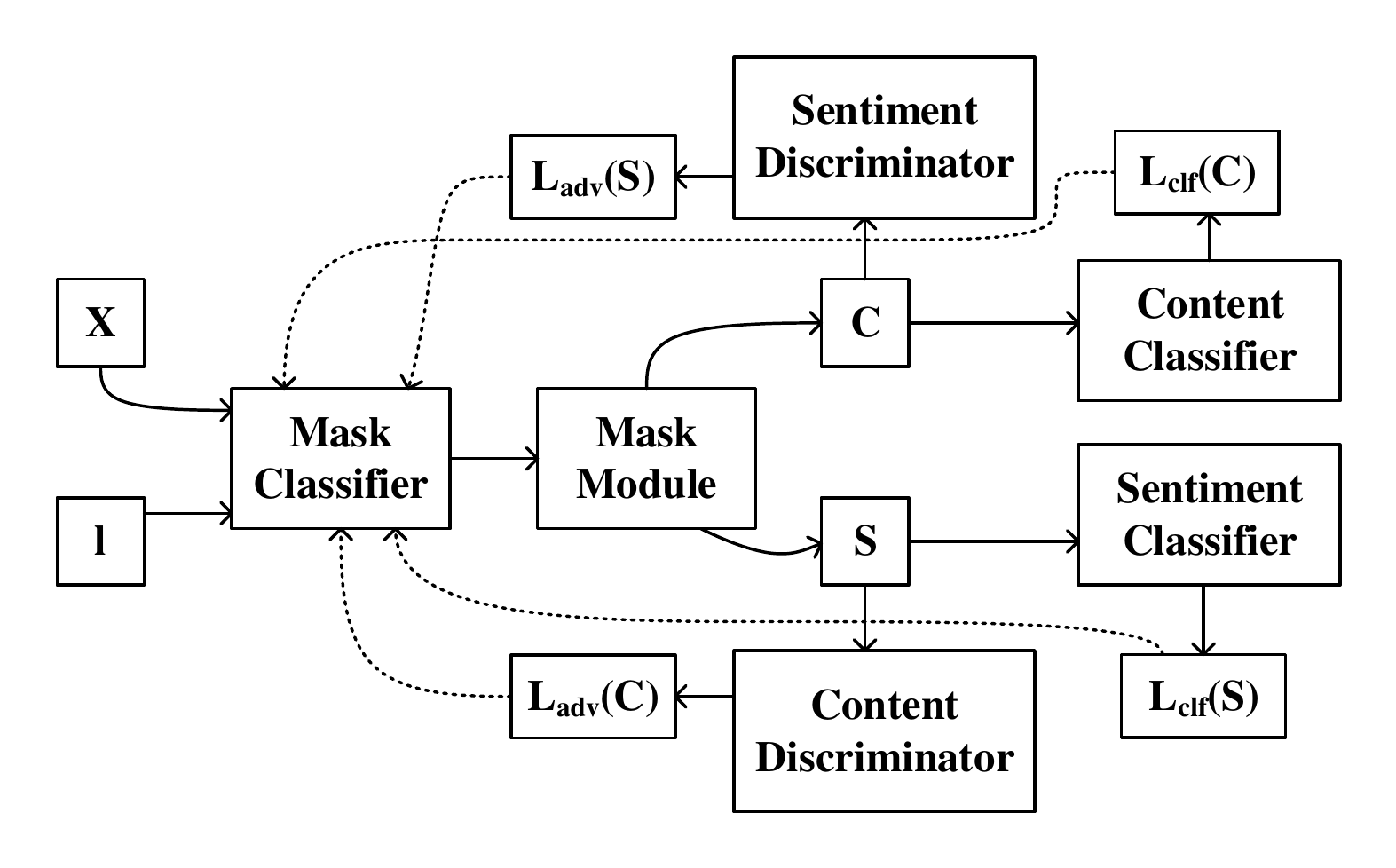}}
\hspace{.1in} 
\subfigure[Infilling Blanks]{
\label{fig2b}
\includegraphics[width=0.47\linewidth]{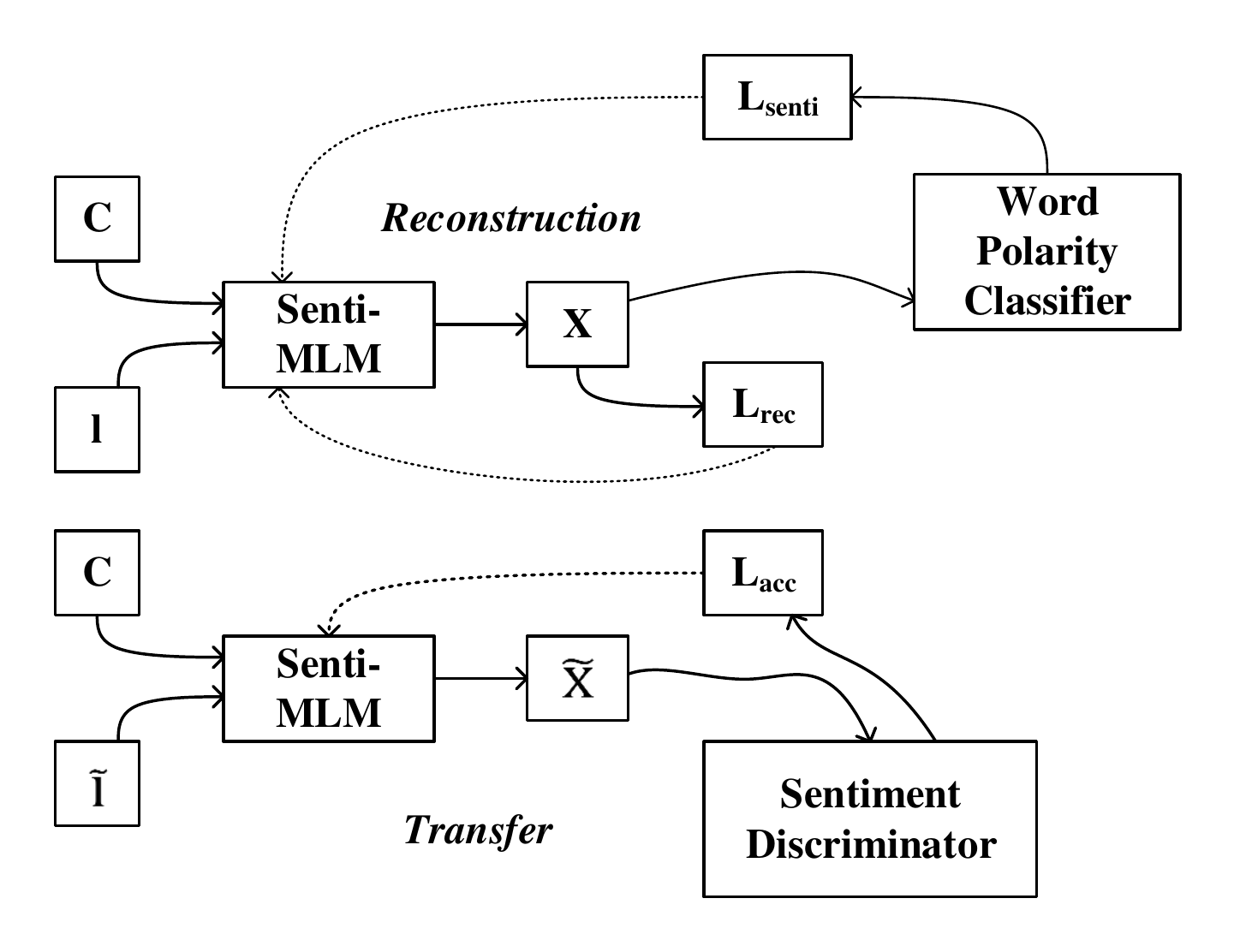}}
\vspace{-4mm}
\caption{Two phases of the proposed AM-ST model.}  
\vspace{-4mm}
\label{fig2} 
\end{figure*}

Given the input text sequence $X$ and the vanilla sentiment label $l$, we first need to decide whether token $x_{i}$ should be masked. A mask classifier as shown in Figure \ref{fig2a} is integrated to learn the masking probability of each token. The mask classifier is implemented as an attention-based classifier.
\begin{equation}
    y=softmax(\alpha \cdot H)
\end{equation}
where $H$ denotes the hidden vectors generated by a Bi-LSTM model \cite{graves2005biLSTM}, and $\alpha$ denotes the attention weight vector generated based on the attention mechanism \cite{vaswani2017attention}. 
The output $y \in \mathbb{R}^{N \times 1}$ indicates the probability of $x_{i}$ containing sentiment information and should be masked. If the predicted probability exceeds a threshold, we will mask the token in this position. We define $S$ as the set of the tokens being masked, and $C$ as the set of the rest of the tokens in $X$. 
The mask classifier will update its parameters adaptively during model training. 
In addition, we introduce extra constraints to facilitate disentangling between $S$ and $C$. Intuitively, all tokens in  $S$ should contain sentiment information but no content information, while tokens in $C$ should only include content information without any sentiment information. 
Such constraints are comprehensively satisfied by the following two types of losses. 




\vspace{-2mm}
\subsubsection{Classification Loss.}
Classification losses ensure that the extracted token set $C$ and $S$ should capture the content and sentiment information, respectively. Thus, we design the following two classification losses for these two types of tokens. 
\vspace{-2mm}
\paragraph{Sentiment Classification Loss.}

The extracted set $S$ should be associated with sentiment information, and we employ a pre-trained sentiment classifier $clf(S)$ implemented as a fully-connected layer with activation function Softmax to examine this. The input of the $clf(S)$ is $S$, and the loss function of $clf(S)$ is: 

\begin{equation}\label{equation2}
     L_{clf}(S)=-\sum_{l \in \text { labels }}  t_{s}(l) \log y_{s(S)}(l)
      \vspace{-1mm}
\end{equation}
where $t_{s} \in \mathbb{R}^{N \times 2}$ denotes the output of $clf(S)$. $y_{s(S)}(l)$ denotes the probability of emotional polarity of $S$ being $l$.
$L_{clf} (S)$ is a cross-entropy loss and approximates the distance between the predicted distribution $y_{s(S)}$ and ground-truth distribution $t_{s}$. 
It is worth noting that classifier $clf(S)$ is pre-trained and its parameters will not be updated during the training of our model, so the reduction of $L_{clf} (S)$ is not caused by the classifier’s better predictive capacity, but brought by richer sentiment information encoded in the input text. 
\vspace{-2mm}
\paragraph{Content Classification Loss.}
To ensure the content information is captured by $C$, we design a content classification loss to measure the closeness between the vanilla input $X$ and the content set $C$ in terms of textual content. Following previous work \cite{john2018}, we use bag-of-words (BoW) to measure content completeness. To testify whether $C$ preserves content well, we employ a Softmax content classifier $clf(C)$ to predict BoW distribution with the input of $C$. The loss function of $clf(C)$ is defined as the cross-entropy between ground-truth distribution $t_{c}$ and predicted distribution $y_{c(C)}$:

\begin{equation}\label{equation3}
  L_{clf}(C)=-\sum_{w \in \text { vocabulary }} t_{c}(w) \log y_{c(C)}(w)
\end{equation} 
where $t_{c}(w)=\frac{\operatorname{count}(w, X)}{N}$ is the ground-truth BoW distribution,
${\operatorname{count}(w, X)}$ is the frequency of word $w$ in the vocabulary appearing in $X$. $y_{c(C)}(w)$ denotes the predicted probability of word $w$’s appearance. 

\vspace{-5mm}
\subsubsection{Adversarial Loss.  }
To separate content and sentiment information, it is indispensable to examine whether $C$ and $S$ contain overlapping information. We further introduce two adversarial losses to accomplish the objective.
\vspace{-3mm}
\paragraph{Sentiment Adversarial Loss.}
To ensure that content set $C$ contains as little sentiment information as possible, we adopt a two-step training paradigm. 
In the first step, we introduce a Softmax sentiment discriminator $dis(S)$ to predict the sentiment label of $C$. To improve the performance of $dis(S)$, we introduce $L_{dis}(S)$ to measure the distance between predicted distribution $y_{s(C)}$ and ground-truth distribution $t_{s}$: 
\begin{equation}
\vspace{-1mm}
  L_{dis}(S)=-\sum_{l \in \text { labels }} t_{s}(l) \log y_{s(C)}(l)
\end{equation} 
where $y_{s(C)}(l)$ represents the  probability of the predicted label of $C$ to be $l$.
As $dis(S)$ has been trained to predict the sentiment label using $C$, in the second step, we introduce $L_{a d v(S)}$ to punish the classification ability of $dis(S)$, which is implemented as the entropy of $y_{s(C)}$ and measures the accuracy of prediction. 
$L_{a d v(S)}$  is maximized when $y_{s(C)}$ is evenly distributed, which means that sentiment labels are completely unpredictable for $C$: 
\begin{equation}\label{equation5}
\vspace{-1mm}
  L_{adv}(S)=-\sum_{l \in \text { labels }} y_{s(C)}(l) \log y_{s(C)}(l)
  \vspace{-1mm}
\end{equation}

\vspace{-5mm}
\paragraph{Content Adversarial Loss.}
In order to remove content information in $S$, we introduce $L_{adv}(C)$, which is calculated in the similar manner of  $L_{adv}(S)$. First, we adopt a content Softmax discriminator $dis(C)$ and optimize its ability to predict content BoW contribution $y_{c(S)}$ using $S$ by minimizing $L_{dis}(C)$.
\begin{equation}
\vspace{-1mm}
  L_{dis}(C)=-\sum_{w \in \text { vocabulary }} t_{c}(w) \log y_{c(S)}(w)
  \vspace{-1mm}
\end{equation}
where $t_{c}$ is the ground-truth distribution and $y_{c(S)}$ is the predicted distribution of $dis(C)$. 
Then, we punish the discernment of content discriminator. $L_{a d v}(C)$ is the entropy of $y_{c(S)}$ and achieves its maximum when it is impossible to discern content distribution using $S$: 
\begin{equation}\label{equation7}
\vspace{-1mm}
  L_{a d v}(C)=-\sum_{w \in \text { vocabulary }} y_{c(S)}(w) \log y_{c(S)}(w)
  \vspace{-1mm}
\end{equation}

\vspace{-5mm}
\subsubsection{Overall objective function} 
Based on the previous losses, the final objective function is formally designed as: 
\begin{equation}\label{equation8}
    \begin{array}{c}
        L_{t o t a l} =\lambda_{1} L_{clf}(S)-\lambda_{2} L_{a d v}(S)+ \lambda_{3} L_{clf}(C)-\lambda_{4} L_{a d v}(C)
    \end{array}
\end{equation}
Where $\lambda _1$,$\lambda _2$,$\lambda _3$ and $\lambda _4$ are the weights of the corresponding losses. 

It is worth noting that although both the attention-based method in \cite{wu2019} and our model utilizes classifiers to find mask position, the former classifier is not updated synchronously during model training, but trained in advance. Therefore, the performance of \cite{wu2019} cannot be continuously improved as it ignores the feedback on whether masked tokens are related to sentiment. Instead, our mask position classifier adaptively updates its parameters using the above four losses and continues improving identifying accuracy.

\vspace{-4mm}
\subsection{Infilling Blanks }

\begin{figure*}[t]
\centering
\includegraphics[width=0.9\textwidth]{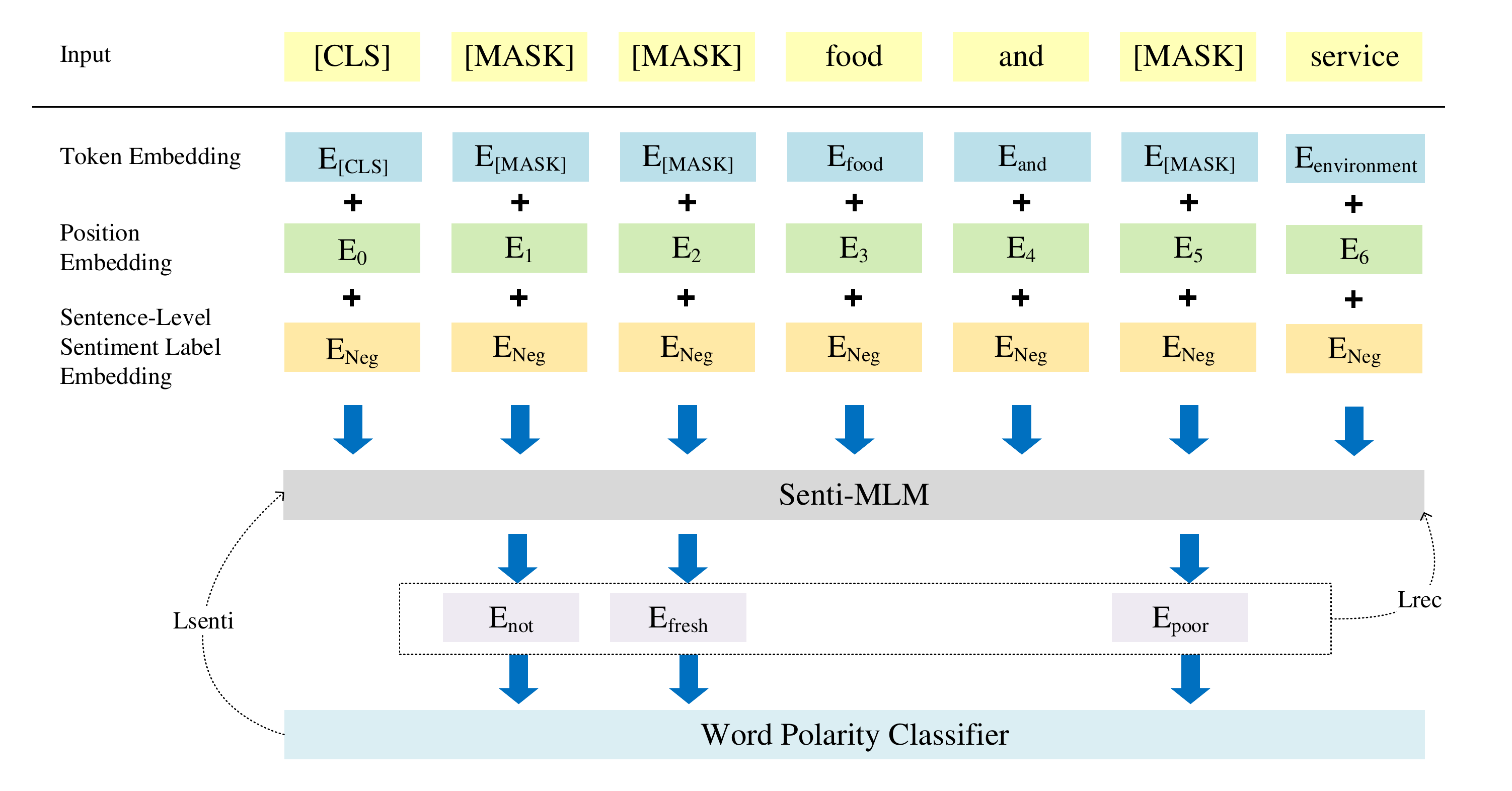} 
\vspace{-4mm}
\caption{Sentiment aware masked language model (Senti-MLM). Token embeddings, position embeddings and sentence-level sentiment embeddings are the inputs, and Senti-MLM is trained to predict tokens and word-level polarities in the masked positions.}
\label{fig3}
\vspace{-4mm}
\end{figure*}

In this stage, our model will infill tokens in masked positions using a sentiment-aware masked language model(Senti-MLM) as shown in Figure \ref{fig2b}. Although MLM performs well in clozing tasks, it only considers contextual information and overlooks sentiment information. However, antonyms with opposite sentiment polarities tend to have different contexts. \cite{xu2020understanding} proved that MLM performs well in learning features of domains and semantics, but tends to neglect opinion words and sentiments. Therefore, simply utilizing MLM to infill words in masked positions may not achieve desirable performance since tokens in these positions still contain rich sentiment information. To solve this problem, we adopt Senti-MLM, which is able to predict the tokens in the masked positions considering not only context but also sentiment information to facilitate transfer accuracy. 

On the basis of the MLM of the pre-trained BERT model, we design a new paradigm to incorporate sentiment information. 
As shown in Figure \ref{fig3}, the traditional segmentation embeddings are replaced by sentence-level sentiment label embeddings. 
The training objective is expected to predict tokens in masked positions and their word-level polarities. 
In the adaptive masking stage, we have proved that all masked tokens are associated with sentiment information. 
Therefore, the proposed Senti-MLM overcomes MLM’s shortcoming of overlooking sentiment information and performs well in using words with proper sentiment polarity to fill in these blanks.

We employ a reconstruction task to train Senti-MLM. Given the set of content tokens $C$ and the original label $l$, the objective aims to reconstruct the input sentence $X$ using Senti-MLM. We measure the performance of the reconstruction task by $L_{rec}$, which reflects the content preservation ability of Senti-MLM: 
\begin{equation}
\vspace{-1mm}
    L_{rec}=\sum_{t_{i} \in S} p\left(t_{i} \mid l ; C\right)
    \vspace{-1mm}
\end{equation}

To take word-level polarities into consideration, we train a word polarity classifier to predict word-level polarities using the hidden-state $h$ generated by Senti-MLM, and its output is $y_{h}$. Then we introduce $L_{senti}$ to measure Senti-MLM’s ability to recover word-level polarities by comparing $y_{h}$ and ground-truth word-level polarity $t_{polar}$.
\begin{equation}
\vspace{-1mm}
    L_{senti}=-\sum_{l \in \text { labels }} t_{polar}(l) \log y_{h}(l)
    \vspace{-1mm}
\end{equation}
Then, loss $L_{r e c}$ and $L_{senti}$ are weighted combined to  finetune Senti-MLM, and $\vartheta_{1}$ and $\vartheta_{2}$ denote the corresponding weights:
\begin{equation}\label{equation 11}
\vspace{-1mm}
    L_{1}=\vartheta_{1} L_{r e c}+\vartheta_{2} L_{senti}
    \vspace{-1mm}
\end{equation}
After that, we leverage Senti-MLM to perform the transfer task. Content tokens $C$ and target sentence label $\widetilde{l}$ are fed into Senti-MLM, and the transferred sentence $\widetilde{X}$ can be generated. To evaluate the transfer accuracy, we introduce $L{acc}$ to measure whether $\widetilde{X}$ conforms with target label $\widetilde{l}$: 
\begin{equation}
    \widetilde{X}=\operatorname{SentiMLM}(C, \widetilde{l})
\end{equation}
\begin{equation}
    L_{a c c}=-\log p(l \mid \widetilde{X})
\end{equation}
We continue finetuning Senti-MLM using $L_{rec}$ and $L_{acc}$. The overall loss in this stage is $L_{2}=\vartheta_{3} L_{r e c}+\vartheta_{4} L_{a c c}$, where $\vartheta_{3}$ and $\vartheta_{4}$ trade off between $L_{r e c}$ and $L_{a c c}$.
\vspace{-4mm}

\section{Experiment}
\vspace{-1mm}

\subsection{Experimental Settings} 
\subsubsection{Dataset}
Following previous works \cite{li2018delete}, we adopt two popular datasets, Yelp\footnote{https://www.yelp.com/dataset} and Amazon\footnote{https://www.kaggle.com/datasets/bittlingmayer/amazonreviews}, to evaluate the performance of our proposal. 
Yelp contains business reviews in which each review is labeled with negative or positive sentiment. 
Similarly, Amazon dataset contains product reviews from Amazon, each of which is manually labeled as either negative or positive. 

\vspace{-2mm}
\subsubsection{Baselines} 

We compare the proposed AM-ST model with the following popular baselines for verifying the performance.
\begin{itemize}
\vspace{-2mm}
    \item CrossAligned~\cite{shen2017}: CrossAligned generates the original sentence back using a generator that combines content representation with the original label.
    \item StyleEmbedding~\cite{fu2018}: Style embedding is fed into a decoder to generate text given different target sentiment signals. 
    \item  MultiDecoder~\cite{fu2018}: MultiDecoder is a  seq2seq model using multiple decoders. Each decoder independently generates a corresponding text style.
    \item CycledReinforce~\cite{xu2018CycledReinforcement}: This model consists of a deemotionalizing module and an emotionalizing module, which extracts non-emotive semantic information and then emotionalizes neutral sentences.
    \item DeleteAndRetrieval~\cite{li2018delete}: DeleteAndRetrieval removes original attribute phrases and retrieves new terms related to the target attribute in the corpus.
    \item DisentangledRepresentation~\cite{john2018}:  VAE with auxiliary multitask and adversarial objectives are used to learn content embeddings and style embeddings.
    \item AC-MLM-Frequency~\cite{wu2019}: AC-MLM-Frequency converts the emotion transfer problem into a clozing task through a masked language model.
    \item AC-MLM-Fusion~\cite{wu2019}:  An extension of AC-MLM-Frequency which employs an attention mechanism to further filter retrieved sentiment words. 
\vspace{-5mm}
\end{itemize}

\begin{table}[t]
\small
\renewcommand\arraystretch{1.1}
\centering
\caption{Dataset statistics.}\label{tab1}
\begin{tabular}{c | c | c c c }
\hline
~~~Dataset~~~ &   ~~~Labels~~~  &  ~~~Train~~~  & ~~Valid~~~ & Test~~~\\
\hline
\multirow{2}*{Yelp } & Positive & 270K & 2000 & 500\\
 &   Negative  &  180K  & 2000 & 500\\
 \hline
\multirow{2}*{Amazon } & Positive & 277K & 985 & 500\\
&   Negative  &  278K  & 1015 & 500\\
\hline
\end{tabular}
\vspace{-2mm}
\end{table}

\subsubsection{Evaluation Metrics}
Following the previous work \cite{li2018delete}, we select Accuracy and BLEU as the evaluation metrics. 
Accuracy is calculated by how likely a transferred sentence conforms with the target sentiment, which is an indicator of transfer accuracy. BLEU is computed by the similarity between human reference by \cite{li2018delete} and the generated transferred sentence. A high BLEU score indicates that the model performs well in content preservation.

\begin{table}[t]
\small
\renewcommand\arraystretch{1.1}
\centering
\caption{Experimental results on Amazon and Yelp. }\label{tab1}
\begin{tabular}{c| c c | c c }
\hline
\multirow{2}*{ } & \multicolumn{2}{c|}{\bfseries Yelp} & \multicolumn{2}{c}{\bfseries Amazon} \\
 &   ~ACC(\%) ~ &  ~BLEU~  & ~ACC(\%)~ & ~BLEU~ \\
\hline
CrossAligned~\cite{shen2017} &  73.1 & 3.1 & 74.1 & 0.4\\
StyleEmbedding~\cite{fu2018}	& 8.7 &	11.8 &	43.3 &	10.0\\ 
MultiDecoder~\cite{fu2018} &	47.6 &	7.1 &	68.3 &	5.0 \\
CycledReinforce~\cite{xu2018CycledReinforcement} &	85.2 &	9.9 &	77.3 &	0.1 \\
DeleteAndRetrieval~\cite{li2018delete} & 88.7 &	8.4 &	48.0 &	22.8 \\
DisentangledRepresentation~\cite{john2018} & 91.5 & 12.2 & 82.4 & 25.2\\
AC-MLM-Frequency~\cite{wu2019} & 95.1 &	11.6 &	64.5 &	27.2 \\
AC-MLM-Fusion~\cite{wu2019} & 95.3 &	12.3 &	85.2 &	28.3 \\
\hline
AM-ST & \bfseries97.1 &	\bfseries 12.9 & \bfseries86.4 &	\bfseries 29.7 \\
\hline
\end{tabular}
\vspace{-4mm}
\end{table}

\vspace{-2mm}
\subsection{Implementation Details}
In the adaptive sentiment token masking phase, we use an attention classifier pre-trained on Yelp and Amazon datasets as the mask classifier. $clf(S)$, $clf(C)$, $dis(S)$ and $dis(C)$ are implemented using four Softmax classifiers.
In the phase of infilling blanks, we use the pre-trained Bert$_{base}$ as the checkpoint. The word polarity classifier is a fully-connected layer with a $768\times3$ tensor as input, and the sentiment discriminator is implemented as a convolutional neural network (CNN). We first finetune the checkpoint to fit Senti-MLM as well as train the word polarity classifier for ten epochs, only $L_{rec}$ is computed in this stage. Then we further train six epochs to finetune Senti-MLM, where both $L_{rec}$ and $L_{senti}$ constitute the total loss in this stage. Finally, we finetune Senti-MLM and train the sentiment discriminator for another ten epochs, where $L_{rec}$ and $L_{acc}$ are used to update the parameters of Senti-MLM and sentiment discriminator. 

As shown in Equation (\ref{equation8}), $\lambda_{1}$, $\lambda_{2}$, $\lambda_{3}$ and $\lambda_{4}$ control the trade-off among four losses in adaptive masking stage. 
Similarly, hyper-parameters  $\vartheta_{1}$, $\vartheta_{2}$, $\vartheta_{3}$ and $\vartheta_{4}$ in Equation (\ref{equation 11}) also define the weights of different losses. 
We carefully tune these hyperparameters on the validation set, and report the testing results of the parameter setting with the best validation performance. 
The tuned value of $\lambda_{1}$, $\lambda_{2}$, $\lambda_{3}$ and $\lambda_{4}$ are 0.2, 0.1, 0.4 and 0.3, respectively.
Best validation performance is achieved when  $\vartheta_{1}$, $\vartheta_{2}$, $\vartheta_{3}$ and $\vartheta_{4}$ are 0.4, 0.2, 0.1 and 0.3. 
\vspace{-3mm}
\subsection{Quantitative Analysis}
\vspace{-1mm}
All sentiment transfer models are evaluated five times, and the average performance is reported in Table \ref{tab1}. 
For the StyleEmbedding model, the parameters of the encoder and decoder are fixed when generating sentences of target sentiments, leading to inferior sentiment capability and thus achieving the worst performance.   
AC-MLMs are the strongest baselines because they take advantage of MLM's strong ability to predict masked tokens according to the semantic context. 
AC-MLM-Fusion achieves better performance since it further overcomes the deficiency of AC-MLM-Frequency by introducing an attention-based classifier to filter pseudo-sentiment words. 
One can clearly see that our proposal consistently outperforms baseline models on both datasets, verifying the effectiveness of the proposed adaptive masking mechanism and the token-level sentiment polarity. 


\begin{table}[t]
\renewcommand\arraystretch{1.1}
\centering
\caption{Ablation Study.}\label{tab3}
\vspace{-2mm}
\begin{tabular}{c | c c | c c }
\hline
\multirow{2}*{ } & \multicolumn{2}{c|}{\bfseries Yelp} & \multicolumn{2}{c}{\bfseries Amazon} \\
 &  ~ ACC(\%) ~ & ~ BLEU ~ & ~ ACC(\%) ~ & ~ BLEU ~\\
\hline
~-$L_{clf}(S)~$	& 96.1 & 12.7 &	84.7 &	29.0\\ 
-$L_{clf}(C)$ &	96.3 &	11.7 &	85.9 &	28.6 \\
-$L_{adv}(S)$ &	96.4 &	12.4 &	85.8 &	29.4 \\
-$L_{adv}(C)$ & 96.8 &	12.1 &	86.2 &	28.9 \\
-$L_{senti}$ & 95.6 & 12.2 & 85.2 & 29.2\\
\hline
AM-ST & \bfseries97.1 &	\bfseries 12.9 & \bfseries86.4 &	\bfseries 29.7 \\
\hline
\end{tabular}
\vspace{-4mm}
\end{table}
\vspace{-2mm}
\subsection{Ablation Study}
\vspace{-2mm}
We further conduct an ablation study to verify the effectiveness of different components. 
Five ablation models are designed by removing different objective functions, namely sentiment classification loss in Equation (\ref{equation2}), content classification loss in Equation (\ref{equation3}), sentiment adversarial loss in Equation (\ref{equation5}) and content adversarial loss in Equation (\ref{equation7}). 
Table \ref{tab3} presents the experimental results of different ablation models. 
One can easily observe that model performance significantly decreases after removing any components, verifying these modules are indispensable to a successful sentiment transfer.  
It is reasonable as the classification losses contribute to capturing the sentiment/content information, while the adversarial losses are capable of separating the sentiment and content information. 
The last module $L_{senti}$ incorporates the token-level sentiment signals into the MLM process, facilitating the generation phase.  
\begin{figure}[h]
\vspace{-4mm}
\centering
\subfigure[ACC vs. $\lambda$ on Yelp]{\label{Fig-4a}
    \resizebox{0.46\columnwidth}{4cm}{%
            \begin{tikzpicture}[font=\Large, scale=0.42]
                \begin{axis}[
                    legend cell align={left},
                    legend style={nodes={scale=1.0, transform shape}},
                    xtick pos=left,
                    tick label style={font=\large},
                    ylabel style={font=\large},
                    ylabel={ACC},
                    xtick={0, 0.1, 0.2, 0.3, 0.4, 0.5},
                    xticklabels={$0$, $0.1$, $0.2$, $0.3$, $0.4$, $0.5$},
                    ytick={96.0, 96.2,96.4,96.6,96.8,97.0, 97.2},
                    yticklabels={$96.0$, $96.2$,$96.4$,$96.6$,$96.8$,$97.0$, $97.2$},
                    legend pos=south east,
                    ymajorgrids=true,
                    grid style=dashed
                ]
                \addplot[
                    color=gray,
                    dotted,
                    mark options={solid},
                    mark=diamond*,
                    line width=1.5pt,
                    mark size=2pt
                    ]
                    coordinates {
                    (0, 96.32)
                    (0.1, 96.88)
                    (0.2, 97.13)
                    (0.3, 96.82)
                    (0.4, 96.75)
                    (0.5, 96.69)
                    };
                    \addlegendentry{$\lambda _1$}
                \addplot[
                    color=purple,
                    dotted,
                    mark options={solid},
                    mark=diamond*,
                    line width=1.5pt,
                    mark size=2pt
                    ]
                    coordinates {
                    (0, 96.92)
                    (0.1, 97.07)
                    (0.2, 96.78)
                    (0.3, 96.76)
                    (0.4, 96.67)
                    (0.5, 96.56)
                    };
                    \addlegendentry{$\lambda _2$}
                \addplot[
                    color=blue,
                    dotted,
                    mark options={solid},
                    mark=*,
                    line width=1.5pt,
                    mark size=2pt
                    ]
                    coordinates {
                    (0, 96.63)
                    (0.1, 96.68)
                    (0.2, 96.70)
                    (0.3, 96.99)
                    (0.4, 97.09)
                    (0.5, 97.05)
                    };
                    \addlegendentry{$\lambda _3$}
                    \addplot[
                    color=teal,
                    dotted,
                    mark options={solid},
                    mark=*,
                    line width=1.5pt,
                    mark size=2pt
                    ]
                    coordinates {
                    (0, 96.09)
                    (0.1, 96.38)
                    (0.2, 96.46)
                    (0.3, 97.11)
                    (0.4, 96.82)
                    (0.5, 96.77)
                    };
                    \addlegendentry{$\lambda _4$}
                \end{axis}
                \end{tikzpicture}
            }
    }
\hfill
\subfigure[ACC vs. $\vartheta$ on Yelp]{\label{Fig-4b}
    \resizebox{0.46\columnwidth}{4cm}{%
            \begin{tikzpicture}[font=\Large,scale=0.42]
                \begin{axis}[
                    legend cell align={left},
                    legend style={nodes={scale=1.0, transform shape}},
                    xtick pos=left,
                    tick label style={font=\large},
                    ylabel style={font=\large},
                    ylabel={ACC},
                    xtick={0, 0.1, 0.2, 0.3, 0.4, 0.5},
                    xticklabels={$0$, $0.1$, $0.2$, $0.3$, $0.4$, $0.5$},
                    ytick={96.0, 96.2,96.4,96.6,96.8,97.0, 97.2},
                    yticklabels={$96.0$, $96.2$,$96.4$,$96.6$,$96.8$,$97.0$, $97.2$},
                    legend pos=south east,
                    ymajorgrids=true,
                    grid style=dashed
                ]
                \addplot[
                    color=gray,
                    dotted,
                    mark options={solid},
                    mark=diamond*,
                    line width=1.5pt,
                    mark size=2pt
                    ]
                    coordinates {
                    (0, 96.12)
                    (0.1, 96.38)
                    (0.2, 96.93)
                    (0.3, 96.99)
                    (0.4, 97.12)
                    (0.5, 97.10)
                    };
                    \addlegendentry{$\vartheta_{1}$}
                \addplot[
                    color=purple,
                    dotted,
                    mark options={solid},
                    mark=diamond*,
                    line width=1.5pt,
                    mark size=2pt
                    ]
                    coordinates {
                    (0, 96.76)
                    (0.1, 96.96)
                    (0.2, 97.12)
                    (0.3, 96.99)
                    (0.4, 96.95)
                    (0.5, 96.89)
                    };
                    \addlegendentry{$\vartheta_{2}$}
                \addplot[
                    color=blue,
                    dotted,
                    mark options={solid},
                    mark=*,
                    line width=1.5pt,
                    mark size=2pt
                    ]
                    coordinates {
                    (0, 96.92)
                    (0.1, 97.08)
                    (0.2, 96.88)
                    (0.3, 96.76)
                    (0.4, 96.67)
                    (0.5, 96.56)
                    
                    };
                    \addlegendentry{$\vartheta_{3}$}
                    \addplot[
                    color=teal,
                    dotted,
                    mark options={solid},
                    mark=*,
                    line width=1.5pt,
                    mark size=2pt
                    ]
                    coordinates {
                    (0, 96.4)
                    (0.1, 96.76)
                    (0.2, 96.82)
                    (0.3, 97.10)
                    (0.4, 96.56)
                    (0.5, 96.56)
                    };
                    \addlegendentry{$\vartheta_{4}$}
                \end{axis}
                \end{tikzpicture}
            }
    }

    \vspace{-3mm}
    \caption{Parameter sensitivity analysis on eight core hyper-parameters.}
    \vspace{-5mm}
\end{figure}
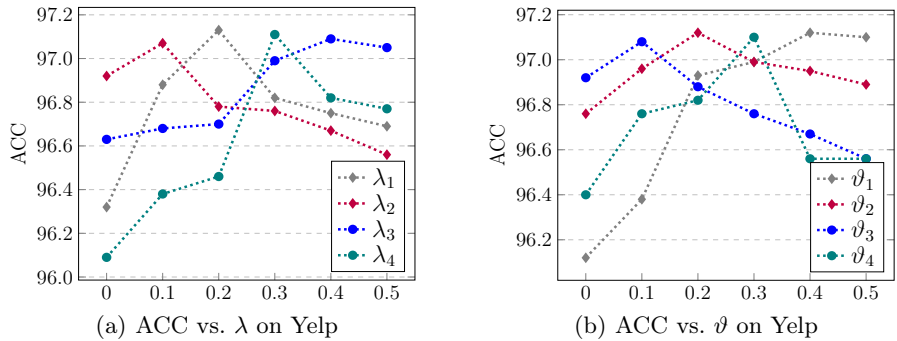
\vspace{-6mm}
\subsection{Parameter Sensitivity Analysis}
\vspace{-1mm}
Here we study the performance sensitivity of our proposal on eight core parameters: the weights $\lambda_{i}$ in Equation (\ref{equation8}) and the weights $\vartheta_{i}$ of Formula (\ref{equation 11}). 
As the performance trends on the two datasets are similar, here we only report the results on the Yelp dataset.   
We first fix other hyper-parameters and then report the results by tuning the target hyper-parameter in the range of $[0, 0.5]$. 
Figure \ref{Fig-4a} and \ref{Fig-4b} presents the experimental results.
One can see that with the increase of different hyper-parameters, the performance over all datasets first increases and then decreases, leading to a similar tendency.  
Thus, these hyper-parameters should be carefully tuned to achieve desirable performance.    
\vspace{-4mm}

\section{Conclusion}
We present a novel model for sentiment transfer, which views the mask positions as trainable parameters to accurately identify and mask sentiment-related words. In addition, a sentiment-aware masked language model is adopted to infill blanks more efficiently by 
considering both context and word-level polarity. Experiments demonstrate that our model consistently outperforms SOTA models.
\vspace{-4mm}

%
%
%
\bibliographystyle{splncs04}

\bibliography{mybib}

\end{document}